\RequirePackage{fix-cm}

\documentclass{article}
\usepackage{cvm21}
\usepackage{amsfonts}
\usepackage{times}
\usepackage{soul}
\usepackage{url}
\usepackage[hidelinks]{hyperref}
\usepackage[utf8]{inputenc}
\usepackage[small]{caption}
\usepackage{graphicx}
\usepackage{amsmath}
\usepackage{amsthm}
\usepackage{booktabs}
\usepackage{algorithm}
\usepackage{algorithmic}
\usepackage{amsmath}
\title{Probability Trajectory: One New Movement Description for Trajectory Prediction}

\author{
	Pei Lv
	\and
	Hui Wei
	\and
	Tianxin Gu
	\and
	Yuzhen Zhang
	\and
	Xiaoheng Jiang
	\and
	Bing Zhou
	\and
	Mingliang Xu
	
	\emails
	(ielvpei,iexhjiang,iebzhou,iexumingliang)@zzu.edu.cn,
	(weihui,txgu,zyzzhang)@gs.zzu.edu.cn
}

\begin{document}

\maketitle

\begin{abstract}
Trajectory prediction is a fundamental and challenging task for numerous applications, such as autonomous driving and intelligent robots. Currently, most of existing works treat the pedestrian trajectory as a series of fixed two-dimensional coordinates. However, in real scenarios, the trajectory often exhibits randomness, and has its own probability distribution. Inspired by this observed fact, also considering other movement characteristics of pedestrians, we propose one simple and intuitive movement description, probability trajectory, which maps the coordinate points of the pedestrian trajectory to a two-dimensional Gaussian distribution in space. Based on this unique description, we develop one novel trajectory prediction method, called \textit{Social Probability}. The method combines the probability trajectory and powerful convolution recurrent neural networks together. Both the input and output of our method are probability trajectories, which provide the recurrent neural network with sufficient spatial and random information of moving pedestrians. Furthermore, the \textit{Social Probability} extracts spatio-temporal features directly on the new movement description to generate robust and accurate predicted results. The experiments on public benchmark datasets show the effectiveness of the proposed method.
\end{abstract}

\section{Introduction}\label{sec:introduction}

The pedestrian trajectory is multimodal, closely related to inherent sense of hearing, vision, touching, thought, personality, and also affected by other factors such as static environments, dynamic human-human interactions, planning destinations, etc. Nevertheless, pedestrians still can intuitively predict the future trajectories of others and adjust themselves in advance. For example, when people walk in shopping malls, streets and stations, they predict the trajectory of others in a short period of time so as to choose their own route at the next moment and avoid collisions. Nowadays, the purpose of trajectory prediction is to enable machines, such as robots, self-driving cars, intelligent tracking systems, to have the ability to predict future trajectories based on historical trajectories. This is one fundamental but extremely challenging task.

\begin{figure}[t]
	\begin{center}
		%\includegraphics[width=0.8\linewidth]{egfigure.eps}
		\includegraphics[width=0.9\linewidth]{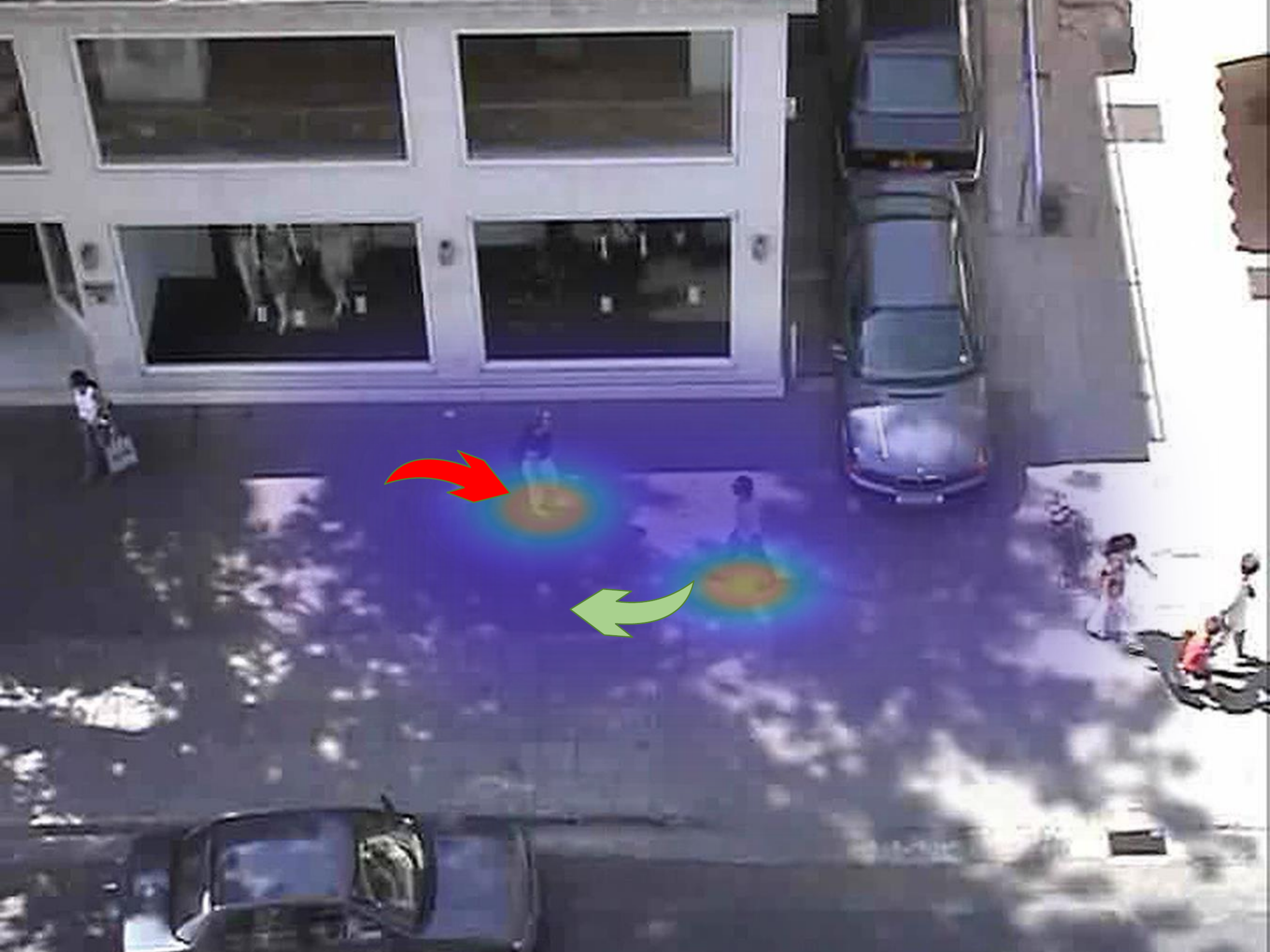}
	\end{center}
	\caption{As illustrated in this scene, two pedestrians are approaching each other. The current position of pedestrians is no longer represented by fixed 2D coordinate points, but a Gaussian probability distribution named after probability trajectory. }
	\label{fig1}
\end{figure}

In previous works, researchers mainly focused on the following problems in trajectory prediction: interaction among pedestrians~\cite{alahi2016social,helbing1995social,vemula2018social,yi2016pedestrian,zhang2019sr,xu2017efficient}, interaction between pedestr-ians and scenes~\cite{liang2019peeking,sadeghian2019sophie,wong2017optimized}, multi-modality~\cite{gupta2018social,liang2020garden}. Especially in recent years, more and more efforts have been made to predict multi-future trajectories~\cite{chai2019multipath,li2019way,makansi2019overcoming,tang2019multiple,xue2018ss} due to the polymorphism of predicted trajectories. In the real world, the trajectory appears as a probability distribution. When the historical trajectory is known and invariant, a person may have many different future trajectories according to dynamic influencing factors. For example, supposing one person walks twice from the same starting location to the same destination, these two trajectories are usually not exactly the same. Although existing works mentioned above were able to predict multi-future trajectories, their inputs were still unimodal, which took the trajectory as two-dimensional coordinate points. Because the movement description is one-dimensional vectors like $({{x}_{t}}, {{y}_{t}})$. Obviously, each coordinate point is invariant and nonrandom. These separate points lose the randomness information of the trajectory. Consequently, these methods cannot fully demonstrate the multi-modality of the trajectory caused by inherent randomness.

In addition, earlier works~\cite{helbing1995social,alahi2016social,vemula2018social,huang2019stgat,sun2020recursive} have made great progress in modeling the impact of human-human interactions. However, there still exists great challenges since most of these works achieve the purpose of modeling pedestrian interactions by combining hidden states. Because the input is a one-dimensional vector, these hidden states are also one-dimensional, which carries little spatial information. The lack of spatial information makes the problem of modeling interactions complicated and incomprehensible.

In order to solve the above problems, we propose the concept of probability trajectory, which is an intuitive and effective motion description. The way of denoting the trajectory is no longer a series of fixed coordinates, but a probability distribution (Fig.~\ref{fig1}). Specifically, we use the probability density function to map the pedestrian coordinate points $({{x}_{t}}, {{y}_{t}})$ to two-dimensional Gaussian distributions $G({{x}_{t}}, {{y}_{t}})$. Compared with fixed coordinate points, the new movement description avoids the loss of randomness. Moreover, we can conveniently map all pedestrians trajectory at time $t$ into one same two-dimensional space. This gives us a unique advantage in modeling human-human interactions.

Based on the proposed probability trajectory, we further develop one new method called \textit{Social Probability} for predicting robust and accurate pedestrian trajectories. First, the input of our method are probability trajectories, which enables our forecasting model fully consider the randomness of the trajectory. Second, by adding convolution layer to recurrent neural network, our forecasting model can learn spatiotemporal features efficiently. We extract the location information of pedestrians from the two-dimensional probability space through the convolutional neural network. Meanwhile, the two-dimensional probability space contains the location of all pedestrians. Through performing convolution operations on the space, we can extract all pedestrians' location information and easily capture changes in relative location. These two factors are indispensable for modeling interaction.

The rest of the paper is organized as follows. In Section~\ref{sec:Relatedwork}, we first review the related work on trajectory prediction. Then we introduce in detail \textit{Social Probability} in Section~\ref{sec:approach}. In Section~\ref{sec:Experiment}, we further present our experimental results as well as the analysis. Finally, we conclude the proposed method and discuss some future directions in Section~\ref{sec:conclusions}.
To summarize, the main contributions of this paper are as follows:
\begin{itemize}
	\item [1.]
	The concept of ``probability trajectory" is proposed to denote pedestrian trajectories, which fully demonstrates the inherent randomness of the trajectory to facilitate the subsequent modeling of its multimodal characteristic.
	\item [2.]
	The \textit{Social Probability} method is further proposed based on the probability trajectory and recurrent neural networks. With the benefit of convolution operation on the probability trajectories, our method can better model human-human interactions in the visual aspect.
	\item [3.]
	The proposed method is successfully tested on public pedestrian datasets. Experiments show that our approach has achieved competitive results both on ADE and FDE compared with other state-of-the-art methods.
\end{itemize}

\section{Related work} \label{sec:Relatedwork}

\subsection{Multimodal trajectory forecasting}
In recent years, some researches have tried to model the randomness of trajectory prediction. Gupta et al.~\cite{gupta2018social} solved the trajectory prediction problem using  Generative Adversarial Networks (GANs)~\cite{goodfellow2014generative} and considered the fact that pedestrian trajectories may have multiple plausible predictions. SoPhie~\cite{sadeghian2019sophie} combined the scene semantic segmentation model with GANs to model trajectories. Multiverse~\cite{liang2020garden} was a jointly model to generate multiple plausible future trajectories, which contained multi-scale location encodings and convolutional RNNs over graphs. Simultaneously, \cite{thiede2019analyzing,tang2019multiple,makansi2019overcoming} also proposed probability networks to solve the problem of the randomness in vehicle trajectory prediction. However, these works all treat position information as two-dimensional coordinate points, and input them into the prediction model, which cannot completely describe the random behavior of pedestrians. Different from these works, we take the probability trajectory transformed from the moving trajectory as input, and generate multimodal future trajectories.

\subsection{Human-human interaction modeling in trajectory forecasting}
For social interaction, researchers utilized multiple methods to establish the interaction model between pedestrians, such as Social Force~\cite{helbing1995social}, Social Pooling~\cite{alahi2016social}, Attention~\cite{vemula2018social}, etc. The methods~\cite{mehran2009abnormal,pellegrini2009you} based on Social Force fully used the principle that attractive forces are used to guide people toward their destinations, and repulsive forces are used to avoid collisions among human-human and human-obstacle. Most of the Social Force-based models try to learn the parameters of the social force functions from real-world crowd datasets. However, Alahi et al.~\cite{alahi2016social} showed the attraction and repulsion alone cannot simulate complex crowd interactions. The approaches~\cite{alahi2016social,gupta2018social,su2017forecast,xu2018encoding} used social pooling layer to allow the LSTMs to share their hidden states. This novel design can model human interaction efficiently, but the complexity will increase when the crowd is dense. Then, the methods based on the attention mechanism have emerged~\cite{sadeghian2019sophie,vemula2018social}. Pedestrains can automatically perceive the importance of certain targets that affects the location at the next time steps. Besides, some methods~\cite{tang2019st,zheng2020gman,yu2017spatio} simultaneously learn spatial and temporal interactive patterns to capture spatio-temporal correlation efficiently and comprehensively. These attempts make the interactive model more suitable for real scenarios. RSBG~\cite{sun2020recursive} established a group-based social interaction model to explore relationships that are not affected by spatial distance, and Graph Convolutional Neural Network~\cite{henaff2015deep} is applied to trajectory prediction. In this paper, the probability trajectory is introduced, and the influence of spatial interaction is automatically perceived through convolution operation, which avoids the design of complex interaction modules. The experimental results show that this method has better interaction performance.

\begin{figure}[t]
	\begin{center}
		%\includegraphics[width=0.8\linewidth]{egfigure.eps}
		\includegraphics[width=1.0\linewidth]{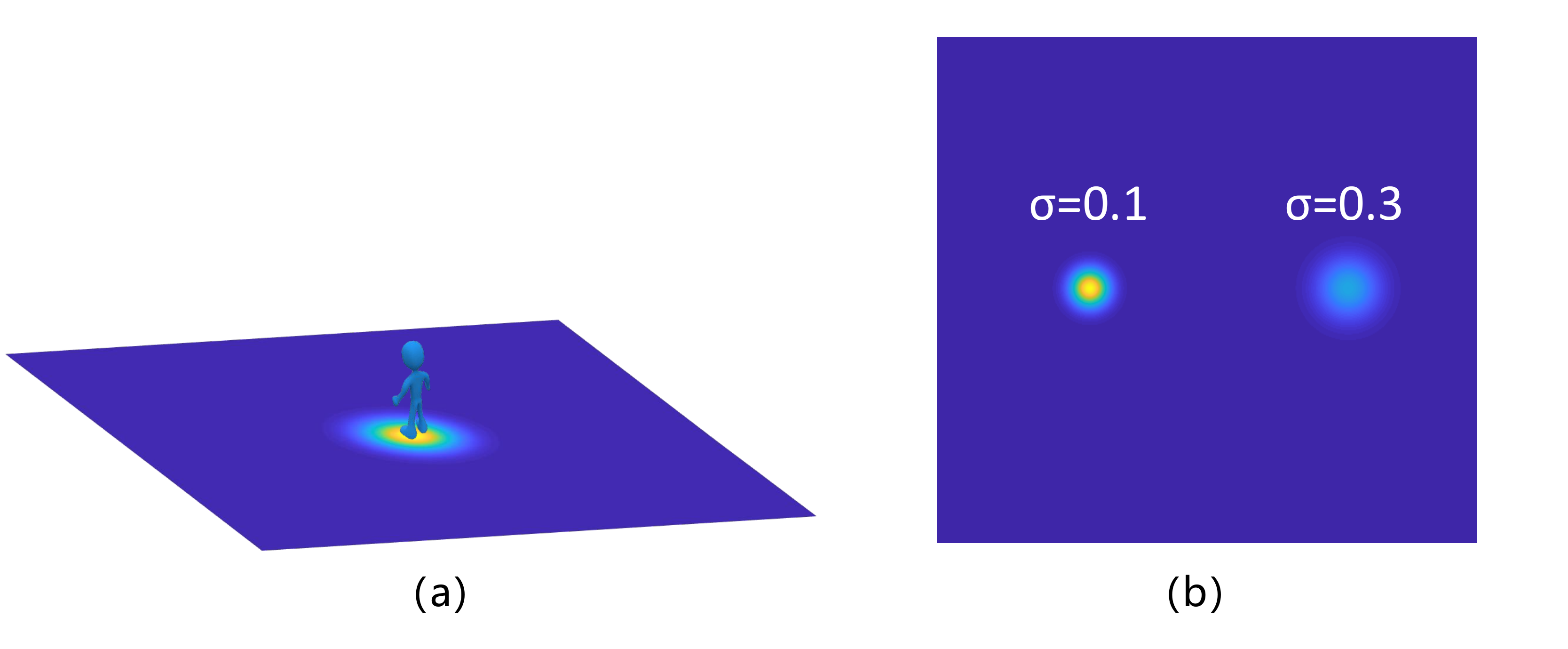}
	\end{center}
	\caption{Probability trajectory describes the location of pedestrians as a probability distribution. (a) shows the distribution of certain trajectory. (b) shows the influential area covered by different~${{\sigma }}$ of the probability distribution. }
	\label{pGraph}
\end{figure}

\subsection{Sequence prediction model}
Sequence prediction is to use the sequences from the past to predict the sequences in the future, which is one kind of time series data modeling problem. Convolutional neural network is very useful in the field of computer vision, but is difficult to learn the characteristics of time series data. The recurrent neural network is specially suitable for dealing with the sequence-related data such as audio, video, text. Recurrent neural network and its derivation LSTM~\cite{hochreiter1997long} and GRU~\cite{cho2014learning} have proved their effectiveness in many fields, such as machine translation~\cite{bahdanau2014neural}, text generation~\cite{karpathy2014deep,vinyals2015show}, speech recognition~\cite{chorowski2014end,chung2015recurrent,graves2014towards}, traffic flow prediction~\cite{yang2019traffic}. Some researchers have combined convolutional neural network with recurrent neural network, and created novel applications, such as image captioning~\cite{vinyals2015show,xu2015show,you2016image}, video understanding~\cite{donahue2015long,srivastava2015unsupervised}. In order to learn spatiotemporal features simultaneously, Shi et al.~\cite{xingjian2015convolutional} have added convolution layer to the recurrent neural network. The model called ConvLSTM not only learns the temporal relationship, but also extracts spatial features by convolution layer. We take the advantages of ConvLSTM to obtain spatiotemporal features and directly model the interaction among pedestrians.

\section{Our approach}\label{sec:approach} 
In this section, we first present the new movement description: probability trajectory, which solves the problem of modeling multimodal trajectories from the data description level, then we propose a prediction model based on probability trajectory to describe human-human interactions conveniently.  
\begin{figure*}[t]
	\begin{center}
		%\includegraphics[width=0.8\linewidth]{egfigure.eps}
		\includegraphics[width=1\linewidth]{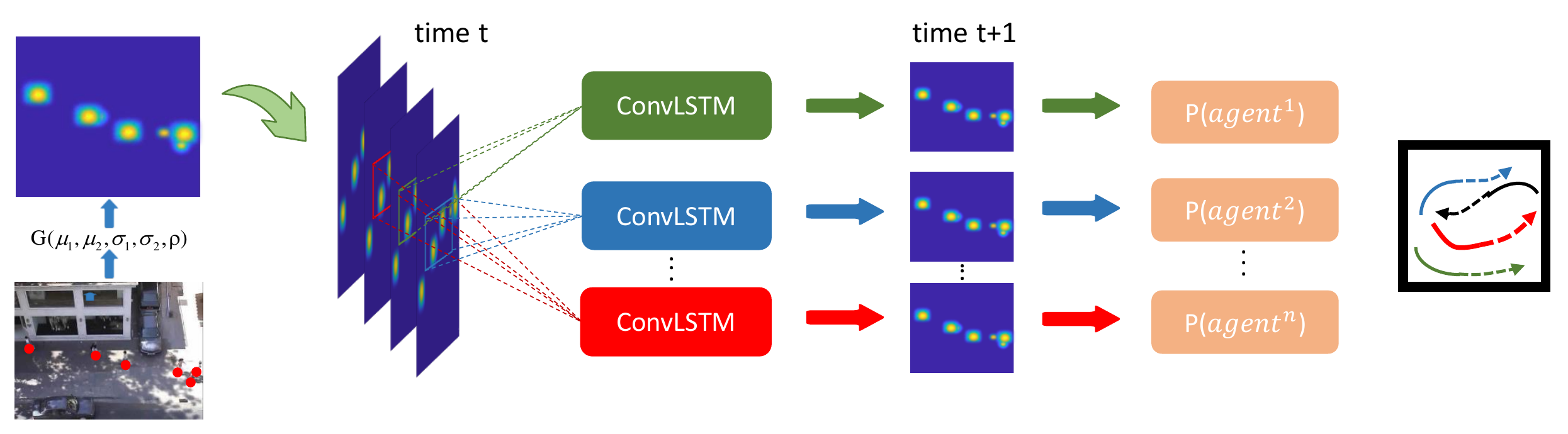}
	\end{center}
	\caption{Overview of our \textit{Social Probability} method. We use a separate ConvLSTM network for each trajectory in a scence. The input of our model is the historical probability trajectories, which are mapped from the fixed coordinate points. One probability trajectory at time $t$ contains the trajectory information of each pedestrian at current time. The ConvLSTM consists of a convolutional layer and gate modules, and has the ability to learn spatiotemporal features. In the predicting stage, the output of our model is also probability trajectories, and trajectory coordinates can be obtained by sampling from them. }
	\label{fig2}
\end{figure*}
\subsection{Problem Definition}
Our goal is to predict the future trajectories of the pedestrians. The input is the historical location information of each pedestrian in the scene and the output is the trajectory information of all people in the future. Define the historical probability trajectory of the pedestrian as $\text{\textbf{X} = }{{\text{X}}_{1}},{{\text{X}}_{2}},......,{{\text{X}}_{n}}$. The predicted future probability trajectory is denoted as $\overset\frown{\textbf{Y}} = \overset\frown{{{Y}_{1}}},\overset\frown{{{Y}_{2}}},......,\overset\frown{{{Y}_{n}}}$, where ${n}$ represents the number of pedestrians. The input trajectory of a pedestrian i is defined as ${X}_{i}\sim N(x_{t}^{i},y_{t}^{i})$ from time steps  $t=1,2,...,{t}_{obs}$ and the future trajectory can be defined similarly as ${Y}_{i}\sim N(x_{t}^{i},y_{t}^{i})$ from time steps  $t={t}_{obs+1},{t}_{obs+2},...,{t}_{pred}$, where $N$ represents Gaussian distribution. The prediction is denoted as $\overset\frown{{Y}_{i}}$ and the ground truth is denoted as ${{Y}_{i}}$.

%-------------------------------------------------------------------------
\subsection{Probability Trajectory}
\label{section:Probability_Trajectory}

\noindent\textit{3.2.1~~Mathematical~definition}

Supposing the feasible area of the pedestrians is $\Omega$, we represent the location of one pedestrian at time $t$ as a probability distribution on $\Omega$. We use the two-dimensional Gaussian distribution, which can well characterize the location of trajectories. The location distribution at time $t$ has the highest probability density at the center position $({{x}_{t}},{{y}_{t}})$. It means that the location does not have to be at this fixed coordinate points and also has a probability of being located in other area. The farther away from the central location point, the smaller the probability density becomes. Note that,  $({{x}_{t}},{{y}_{t}})$ obeys the two-dimensional Gaussian distribution with parameters $({{\mu }_{1}},{{\mu }_{2}},{{\sigma }_{1}},{{\sigma }_{\text{2}}},\rho )$:
\begin{equation}
	\begin{split}
		f({{x}_{t}},&{{y}_{t}})={{(2\pi {{\sigma }_{1}}{{\sigma }_{2}}\sqrt{1-{{\rho }^{2}}})}^{-1}}\exp [-\frac{1}{2(1-{{\rho }^{2}})} \\
		&(\frac{{{({{x}_{t}}-{{\mu }_{1}})}^{2}}}{{{\sigma }_{1}}^{2}}-\frac{2\rho ({{x}_{t}}-{{\mu }_{1}})({{y}_{t}}-{{\mu }_{2}})}{{{\sigma }_{1}}{{\sigma }_{2}}})+\frac{{{({{y}_{t}}-{{\mu }_{2}})}^{2}}}{{{\sigma }_{2}}^{2}}]
	\end{split}
\end{equation}
where ${{\mu }_{1}}$ and ${{\mu }_{2}}$ are the mean value of $({{x}_{t}},{{y}_{t}})$ respectively, ${{\sigma }_{1}}$ and ${{\sigma }_{2}}$ are the variance of $({{x}_{t}},{{y}_{t}})$, $\rho $ is the correlation coefficient of ${{x}_{t}}$ and ${{y}_{t}}$. ${{\mu }_{1}}$ is set to ${{x}_{t}}$ and ${{\mu }_{2}}$ is set to ${{y}_{t}}$. ${{\sigma }_{1}}$ and ${{\sigma }_{2}}$ are set to 0.3 according to the experience, and $\rho $ is 0.

Using this data structure to represent trajectories, we successfully retain the randomness of trajectories. In two-dimensional space, a pedestrian trajectory is no longer a single point at time $t$, but a probability distribution. As shown in Fig.~\ref{pGraph}(a). 

\noindent\textit{3.2.2~~Integrate neighbor's information}

At time $t$, we denote the probability trajectory of pedestrian $i$ as $pMap_{t}^{i}$. However, the scene at time $t$ contains multiple pedestrians. Neighboring pedestrians have great influence on the movement decision of each subject pedestrian. In order to enable the model to predict future trajectories based on the location information of the surrounding pedestrians, we need to integrate the probability trajectories of all pedestrians at time $t$ into a same two-dimensional probability space. The probability trajectory at time $t$ denotes $pMap_{t}$. In two-dimensional space, we integrate $pMap_{t}^{i}$ into $pMap_{t}$ by $\max (\cdot )$ function. Specifically, for the corresponding position in the probability trajectory, we take the larger value as the consolidated value. The formula is as follows. 
\begin{equation}
	\begin{split}
		pMa{{p}_{t}}=max (pMa{{p}_{t}}, pMap_{t}^{i}), i=1,2,...,n
	\end{split}
\end{equation}
where $n$ is the number of pedestrians at time $t$. In order to distinguish the current predicted pedestrian from the rest of the surrounding pedestrians, we set the ${\sigma }$ values to 0.1 and 0.3 respectively. The comparison of different ${\sigma }$ is shown in Fig.~\ref{pGraph}(b). 
%-------------------------------------------------------------------------
\subsection{Convolutional LSTM}
Due to its unique structure, the long and short-term memory network (LSTM) has great advantages in processing time sequence data. Moreover, Shi et al.~\cite{xingjian2015convolutional} proposed variant of LSTM, which added the convolutional layer to the LSTM module, called ConvLSTM, and proved that the model can learn spatio-temporal information through experiments. Specifically, the main operations are as follows:
\begin{align}
	& {{i}_{t}}=\sigma ({{W}_{xi}}{{X}_{t}}+{{W}_{hi}}{{h}_{t-1}}+{{W}_{ci}}{}^\circ {{c}_{t-1}}+{{b}_{i}}) \\
	& {{f}_{t}}=\sigma ({{W}_{xf}}{{X}_{t}}+{{W}_{hf}}{{h}_{t-1}}+{{W}_{cf}}{}^\circ {{c}_{t-1}}+{{b}_{f}}) \\
	& {{c}_{t}}={{f}_{t}}{}^\circ {{c}_{t-1}}+{{i}_{t}}{}^\circ \tanh ({{W}_{xc}}{{X}_{t}}+{{W}_{hc}}{{h}_{t-1}}+{{b}_{c}}) \\
	& {{o}_{t}}=\sigma ({{W}_{xo}}{{X}_{t}}+{{W}_{ho}}{{h}_{t-1}}+{{W}_{co}}{}^\circ {{c}_{t}}+{{b}_{o}}) \\
	& {{h}_{t}}={{o}_{t}}{}^\circ \tanh ({{c}_{t}})
\end{align}
where ${{X}_{t}}$ is the input of time $t$, ${{h}_{t}}$ and ${{c}_{t}}$ are hidden state and cell state, respectively. ${{i}_{t}}$, ${{f}_{t}}$, ${{o}_{t}}$ are the gates of the ConvLSTM. They are all 3-dimensional tensors whose last two dimensions are spatial dimensions (width, height). $W$ is the weight matrix. ‘◦’ denotes the Hadamard product. At time $t$, ${{X}_{t}}$ can be input to the module for calculation only when the input gate is activated. Similarly, the past cell state ${{c}_{t-1}}$ will be forgotten when the forget gate ${{f}_{t}}$ is activated and the current cell state ${{c}_{t}}$ will be transfered when the output gate ${{o}_{t}}$ is on.

The ConvLSTM uses the current input and past states to determine the future states, and the current input includes not only temporal features, but also spatial features. The temporal features can be learned through the gate structure mentioned above, and the spatial features can be extracted through the convolutional layer embedded in the module. Essentially, trajectory prediction can be regarded as a spatiotemporal sequence generation problem. Therefore, applying ConvLSTM to solve it, we can model the temporal characteristics of the trajectory while also considering the spatial interaction of different trajectories.
%-------------------------------------------------------------------------

\subsection{Social Probability}
As illustrated in Fig.~\ref{fig2}, the \textit{Social Probability} is one trajectory prediction method based on probability trajectories. Firstly, we map the position information of all pedestrians at time $t$ into probability trajectories. Then, the ConvLSTM module takes two-dimensional probability trajectories as input and outputs predictive probability trajectories at future time $t+1$. The coordinate points of trajectories can be obtained by sampling from outputs.

\noindent\textit{3.4.1~~Probability-based prediction}

The input to the ConvLSTM needs to be two-dimensional tensors. As discussed in Section~\ref{section:Probability_Trajectory}, our probability trajectory is a probability distribution in two-dimensional space. Therefore, it is suitable to input probability trajectory into the ConvLSTM model. Moreover, probability trajectories are essent-ially probability density distributions. The value of the probability trajectory indicates the level of probability density. Modeling probability trajectories directly makes our method a probability-based forecasting method. Our method not only predicts the multimodal future trajectory, but the input historical trajectory is also multimodal, which is different from previous methods. The problem of modeling multimodal fea-tures is solved from the data level.

\noindent\textit{3.4.2  Human-human interactions modeling }

\begin{figure}[t]
	\begin{center}
		%\includegraphics[width=0.8\linewidth]{egfigure.eps}
		\includegraphics[width=1.0\linewidth]{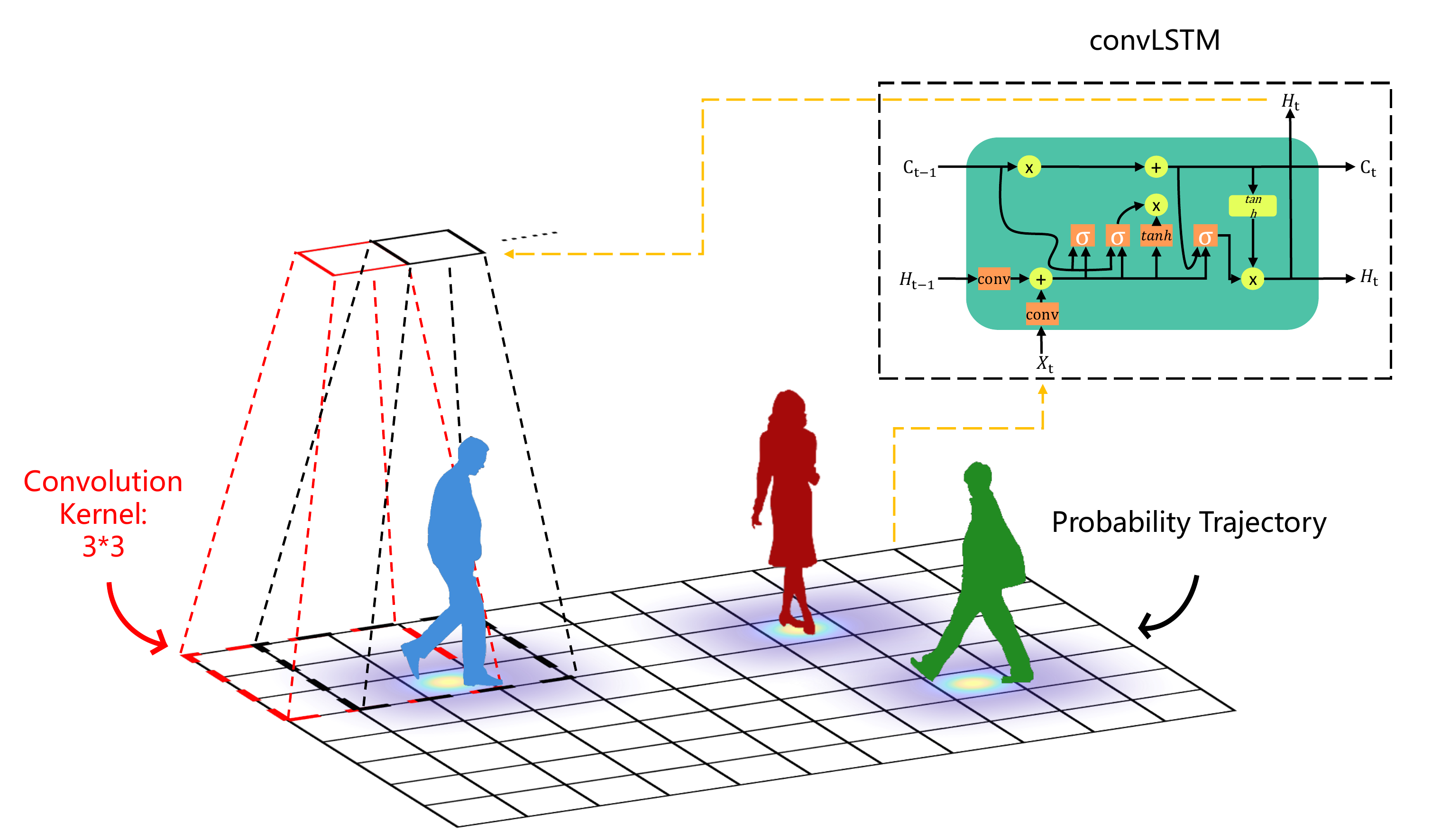}
	\end{center}
	\caption{The convolution layer added into LSTM makes the model have the ability to extract spatial features. Convolution operation on the probability trajectory enables our method to model human-human interaction more intuitively and efficiently.}
	\label{convfig}
\end{figure}

The input of our model is probability trajectories of all pedestrians at time $t$, and is integrated into one two-dimensional space, so modeling human-human interactions is direct and expediently. As illustrated in Fig.~\ref{convfig}, after probability trajectories are input into the model, the convolutional layer will extract features in the two-dimensional probability trajectory to obtain the hidden state, which is the feature vector in the RNN-based model. Since the convolution kernel slides across the entire two-dimensional space like a sliding window, hidden states contain the location information of each pedestrian. Namely, due to the convolution operation, the model not only considers the density value of the current position, but also the density value of the surrounding positions when predicting the probability density value at the next time. Therefore, our model considers the location information of all pedestrians at time $t$, which promotes human-human interactions without complex interaction modules.

\noindent\textit{3.4.3  Loss function for probability trajectory}

We empirically choose the loss function to train our model by refering to the previous works~\cite{feng2018wing,zhao2016loss}. Since our model focuses on the specific probability density value, rather than some high-dimensional features, such as style, graphics, objects, we use $L2$ loss function to encourage our model to generate accurate probability density distributions.
\begin{equation}
	\begin{split}
		{{\mathcal L}_{L2}}(\hat{Y}_{t}^{i},Y_{t}^{i})=||\hat{Y}_{t}^{i}-Y_{t}^{i}|{{|}^{2}}
	\end{split}
\end{equation}
Here, $\hat{Y}_{t}^{i}$ and $Y_{t}^{i}$ are predicted and ground truth probability trajectory for person $i$ at time $t$ respectively.

\begin{table*}
	\begin{center}
		\begin{tabular}{l||c|c|c|c|c||c}
			\hline
			Method & ETH & HOTEL & UNIV & ZARA1 & ZARA2 & AVG \\
			\hline\hline
			Linear & 1.33 / 2.94 & 0.39 / 0.72& 0.82 / 1.59 & 0.62 / 1.21 & 0.77 / 1.48 & 0.79 / 1.59\\
			Vanilla-LSTM & 1.09 / 2.14 & 0.86 / 1.91 & 0.61 / 1.31 & 0.41 / 0.88 & 0.52 / 1.11 & 0.70 / 1.52\\
			Social-LSTM~\cite{alahi2016social} & 1.09 / 2.35 & 0.79 / 1.76 & 0.67 / 1.40 & 0.47 / 1.00 & 0.56 / 1.17 & 0.72 / 1.54 \\
			Social-GAN~\cite{gupta2018social} & 0.81 / 1.52 & 0.72 / 1.61 & 0.60 / 1.26 & 0.34 / 0.69 & 0.42 / 0.84 & 0.58 / 1.18 \\
			Social-GAN-P~\cite{gupta2018social}& 0.87 / 1.62 & 0.67 / 1.37 & 0.76 / 1.52 & 0.35 / 0.68 & 0.42 / 0.84 & 0.61 / 1.21 \\
			SoPhie~\cite{sadeghian2019sophie} & \textbf{0.70} / 1.43 & 0.76 / 1.67 & \textbf{0.54} / 1.24 &\textbf{0.30 / 0.63} & 0.38 / 0.78 & 0.54 / 1.15 \\
			RSBG~\cite{sun2020recursive} & 0.80 / 1.53 & 0.33 / 0.64 & 0.59 / 1.25 & 0.40 / 0.86 & \textbf{0.30 / 0.65} & 0.48 / 0.99 \\
			NEXT~\cite{liang2019peeking} & 0.73 / 1.65 & \textbf{0.30 / 0.59} & 0.60 / 1.27 & 0.38 / 0.81 & 0.31 / 0.68 & \textbf{0.46} / 1.00 \\
			\hline
			Ours & 0.74 / \textbf{1.22} & 0.49 / 0.85 & 0.63 / \textbf{1.23} & 0.42 / 0.78 & 0.38 / 0.70 & 0.53 / \textbf{0.95} \\
			\hline
		\end{tabular}
	\end{center}
	\caption{ Quantitative results of different methods on ETH, Hotel (from ETH) and UNIV, ZARA1, ZARA2 (from UCY) datasets. We show two error metrics Average Displacement Error (ADE) and Final Displacement Error (FDE) on the task of predicting 12 future time steps.}
	\label{table1}  
\end{table*}

\section{Experiment}\label{sec:Experiment}

In this section, we show the experimental results on five public datasets, and compare them with current state-of-the-art methods, then analyze the performance of our method.
\subsection{Datasets}
We validate the proposed model on the public datasets ETH~\cite{pellegrini2009you} and UCY~\cite{lerner2007crowds}, which are the widely used benchmark datasets in the field of trajectory prediction. Most of the current state-of-the-art meth-ods are evaluated on these datasets. It contains a total of 1536 labeled pedestrians in 4 different scenes. These datasets are based on binocular vision for the research of pedestrian trajectory tracking and prediction. There are totally 5 sub datasets, where ETH contains two sub datasets as ETH and HOTEL, and UCY consist of three parts as ZARA1, ZARA2 and UNIV. Similar to the previous works, we still observe the historical trajectory for the past 8 time steps (3.2 seconds) and predict the future trajectory for the next 12 time steps (4.8 seconds).
\subsection{Evaluation Metrics and Methods}
According to previous works~\cite{alahi2016social}, we use two evaluation metrics. 
\begin{itemize}
	\item [1.] 
	\textit{Average displacement error (ADE):} The average Euclidean distance between the predicted trajectories and the true trajectories at each prediction time step.
	\item [2.] 
	\textit{Final displacement error (FDE):} The Euclidean distance between the predicted destination and the ground truth destination at the last prediction time step.
\end{itemize}
The two evaluation scales are defined as:

\begin{align}
	& ADE=\frac{\sum\limits_{i\in \mathbb{Z}}{\sum\limits_{t={{T}_{obs}}+1}^{{{T}_{pre}}}{\sqrt{{{((\hat{x}_{t}^{i},\hat{y}_{t}^{i})-(x_{t}^{i},y_{t}^{i}))}^{2}}}}}}{\mathbb{Z}*{{T}_{pre}}} \\
	& FDE=\frac{\sum\limits_{i\in \mathbb{Z}}{\sqrt{{{((\hat{x}_{{{T}_{pre}}}^{i},\hat{y}_{{{T}_{pre}}}^{i})-(x_{{{T}_{pre}}}^{i},y_{{{T}_{pre}}}^{i}))}^{2}}}}}{\mathbb{Z}}
\end{align}
where $(\hat{x}_{t}^{i},\hat{y}_{t}^{i})$ and $(x_{t}^{i},y_{t}^{i})$ are the predited and ground truth coordinates for pedestrian $i$ at time $t$, $\mathbb{Z}$ is the total number of pedestrian in the testing set.

We use a leave-one-out approach to evaluate the performance of the model. Four sets are used as the training set and verification set, and the remaining one is used as the testing set to gain the results.

\subsection{Implementation Details}
The number of layers of the ConvLSTM model is 5 and the channel dimension of the hidden state in each layer is 128, 64, 64, 32, 32 respectively. The kernel size of the convolutional layer is 3*3 and the padding is 1. We train our model using Adam~\cite{kingma2014adam} with the initial learning rate of 0.001. The size of the probability trajectory and the hidden state of our model are both 100*100. In the stage of prediction, the variance of the current pedestrian to be predicted is set to 0.1, and the other pedestrians are set to 0.3. In the testing stage, we sample 20 times from the probability trajectory predicted by the model, and select the best prediction in Euclidean distance for quantitative estimation.

\subsection{Compared With different Methods}
As shown in Table~\ref{table1}, we choose the following methods for comparison:
\begin{itemize}
	\item [1.] 
	Linear: A linear regression model to predict the trajectory by minimizing the least square error.
	\item [2.] 
	Vanilla-LSTM: Use the LSTM model to predict the future trajectory. This method only considers its own historical trajectory and does not consider any other factors.
	\item [3.] 
	Social-LSTM~\cite{alahi2016social}: The social-pooling layer is added to LSTM, so that the model has the ability to model human-human interactions.
	\item [4.] 
	Social-GAN~\cite{gupta2018social}: A trajectory prediction model trained with GAN architecture is designed to improve existing models in terms of rationality, diversity, and prediction speed. The model pays attention to the feasibility of predictive generation trajectory in social rules.
	\item [5.] 
	Social-GAN-P~\cite{gupta2018social}: The only difference with the Social-GAN is that the pooling mechanism is not applied.
	\item [6.] 
	SoPhie~\cite{sadeghian2019sophie}: An interpretable framework based on GAN for trajectory prediction. It uses two information sources, the historical trajectory of all pedestrians in a scene and the scene context information of the scene image.
	\item [7.] 
	RSBG~\cite{sun2020recursive}: A group-based social interaction model to explore pedestrian relationships that are not affected by spatial distance. Graph Convolutional Neural Network is applied to trajectory prediction in this model.
\end{itemize}

\begin{figure*}[t]
	\begin{center}
		%\includegraphics[width=0.8\linewidth]{egfigure.eps}
		\includegraphics[width=1.0\linewidth]{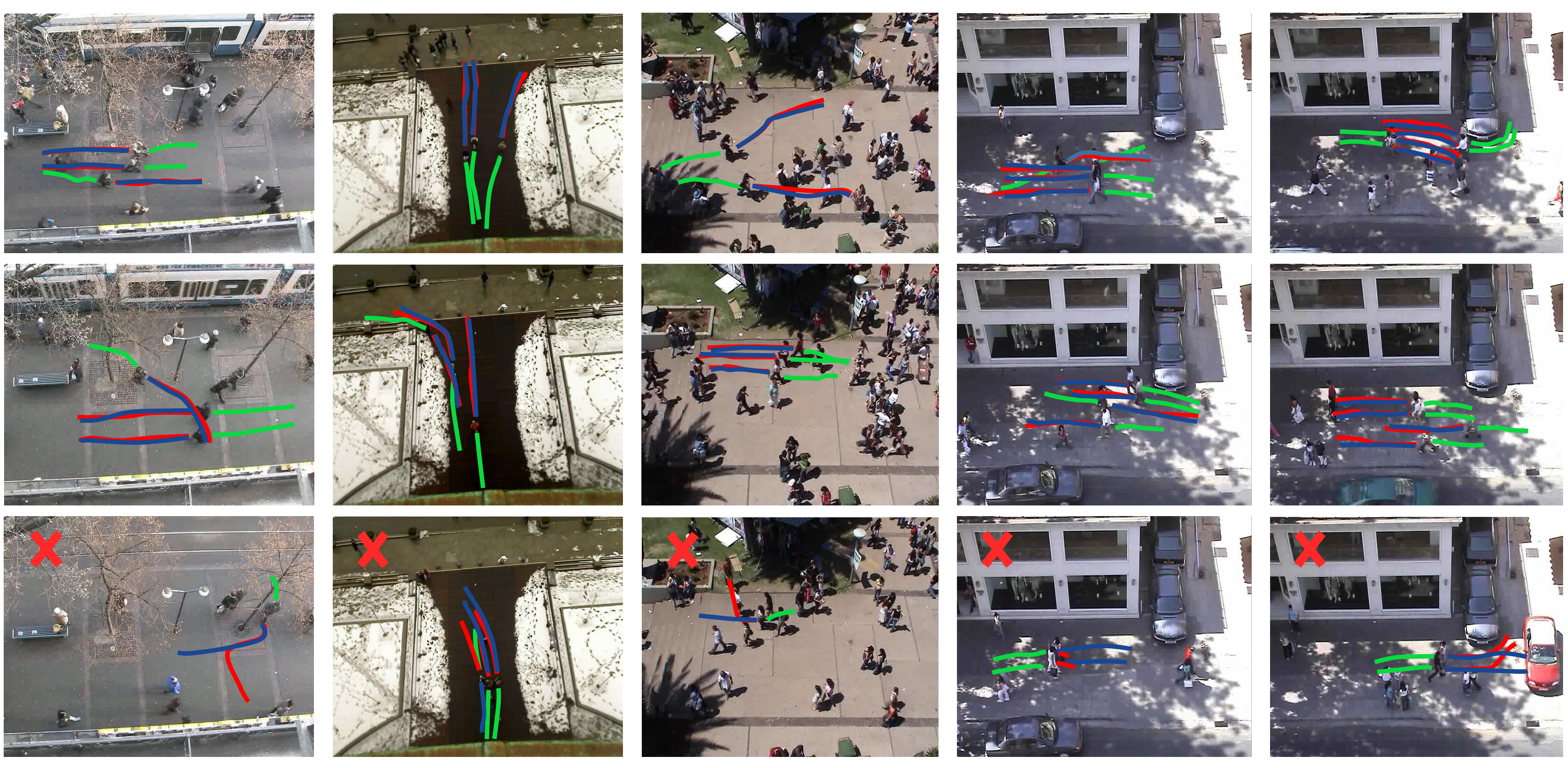}
	\end{center}
	\caption{Visualization of qualitative analysis. Given the observed trajectories of past eight time steps(Green), our method predicts the trajectories of future twelve time steps(Blue), and the ground truth trajectories(Red) is compared with the predicted trajectory. The results for ETH, HOTEL, UNIV and ZARA1 and ZARA2 are shown in column one to five, respectively. On the first two row, we show examples where our method successfully predicts the trjectories with small errors. The last row shows some failure cases. Note that, these   are representative examples selected from the visual results.  }
	\label{fig6}
\end{figure*}

\subsection{Quantitative Analysis}
Table~\ref{table1} lists two error metrics Average Displacement Error (ADE) and Final Displacement Error (FDE) of our method against existing methods, with the task of predicting 12 future time steps according to 8 historical time steps. We follow the comparative works to choose the best prediction among multiple samples for quantitative analysis. It can be seen that the linear model usually performs the worst. Because it is only suitable to predict the straight trajectories, and is insensitive to pedestrian interaction. Social-LSTM and Social-GAN perform better than Linear method since they can handle interacrions among pedestrians by introducing the corresponding interactive module. We can see that our method outperfroms all others on FDE of the ETH and UNIV datasets, avoiding more potential collisions in the future. Although the performance of our method is not the best on other datasets, they are still very competitive and significantly superior to the linear model except for the HOTEL dataset with a small amount of pedestrians. Therefore, it fully proves that our method own inherent interaction function without complex interaction modules. In addition, compared with ADE, our method performs better in FDE, especially in the ETH dataset. This reflects that our method has more advantages in the aspect of predicting destinations.
\begin{table}
	\begin{center}
		\begin{tabular}{l||c|c}
			\hline
			Method & ADE & FDE  \\
			\hline
			Our method with attention & 0.99 & 1.73 \\
			\hline
			Our method without attention & 0.91 & 1.61 \\
			\hline
		\end{tabular}
	\end{center}
	\caption{ Ablation study for attention mechanism. We show the influences when adding attention to our methods. The experiment was conducted on ETH dataset without data augmentation.}
	\label{table2}  
\end{table}

\subsection{Qualitative Analysis}
Fig.~\ref{fig6} shows the positive and negative samples on each dataset. The blue trajectories are the predictive trajectories of future 12 time steps from the observed trajectories of past 8 time steps marked as green, and the red trajectories are the ground truth trajectories. The scenarios in the ETH, HOTEL, UNIV, ZARA1 and ZARA2 dataset are shown in column 1 to 5. The visualization results show that our model is able to correctly predict the future path and have the ability to model human-human interactions. According to the positive samples, the model can avoid obstacles in advance when interacting with others. Besides, our method is also suitable for crowded scenes, when multiple people are walking forward to the same or different direction, voiding each other by following or interpolation.

The last row shows some negative samples. These examples have large gaps in predictions, or have the wrong direction. By analyzing the source videos, we found that these failure cases were generated when pedestrians stopped walking or turned suddenly. The main reason for these results is the unpredictability of pedestrian intentions. Another reason is that when pedestrians interact with the physical surrounding, the model cannot well perceive the scene information. Our method has not yet covered the integration of scene information which is the direction of our future research.

\subsection{Ablation Study}
\noindent\textit{4.7.1  Attention mechanism}

During the experiment, we try to use spatial attention mechanism~\cite{woo2018cbam} to improve the prediction accuracy of our model. In the two-dimension space of probability trajectory, attention module is applied to capture which location has more influence. However, we found that the attention mechanism did not improve our experimental results as expected. The comparative experiments are shown in Table~\ref{table2}. The reason may be that the probability trajectory has already played a role in attention. The probability density of each spatial position represents the importance of the location, namely the weight value in the attention mechanism.

\begin{table}
	\begin{center}
		\begin{tabular}{l||c|c}
			\hline
			Method & ADE & FDE  \\
			\hline
			Our method (full algorithm) & 0.74 & 1.22 \\
			\hline
			Our method without integration & 0.86 & 1.64 \\
			\hline
		\end{tabular}
	\end{center}
	\caption{ Ablation study for our full method without the integration of neighbor's probability trajectories. The experiment was conducted on ETH dataset. }
	\label{attention}  
\end{table}

\noindent\textit{4.7.2  The integration of probability trajectories}

As discussed in Section~\ref{section:Probability_Trajectory}, we integrate the probability trajectories of all pedestrians at time $t$ into one same two-dimensional probability space. In this section, we remove the integration to verify the ability to model interaction of our method. When predicting the trajectory of the person $i$, the probability trajectory only contains its own trajectory information. The trajectory information of people around is not integrated into his probability trajectory. We conduct experiments on the ETH dataset and the results are shown in Table~\ref{attention}. We can find that the method with integration reaches an improvement of 14.0 $\%$ and 25.6 $\%$ in ADE and FDE. It is fully proved that the integration of the probability trajectories has the ability to model human-human interactions.

\begin{figure}[t]
	\begin{center}
		%\includegraphics[width=0.8\linewidth]{egfigure.eps}
		\includegraphics[width=1.0\linewidth]{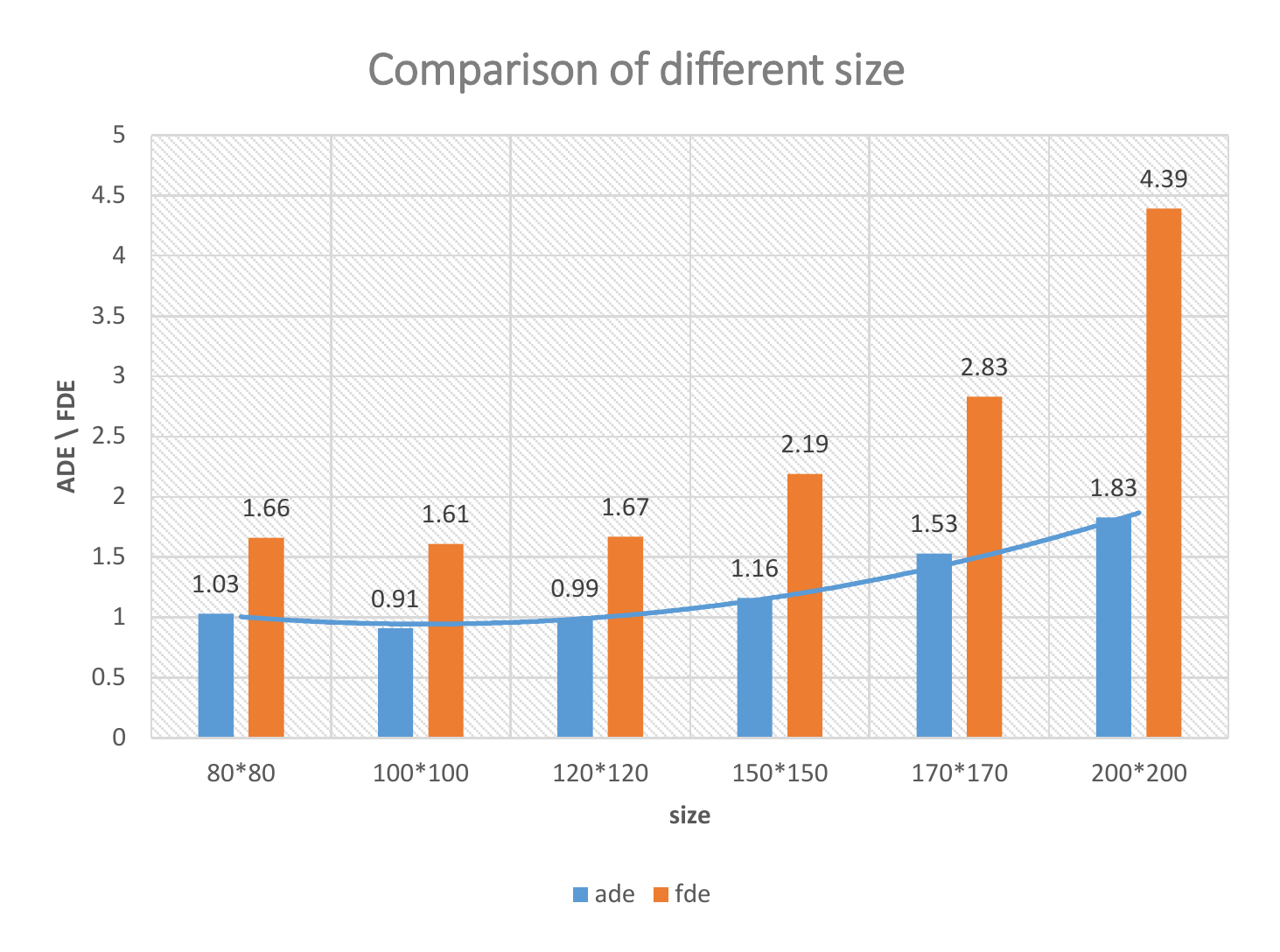}
	\end{center}
	\caption{Ablation study for different sizes of probability trajectories. The blue line shows the trend of ADE. Experiments was conducted on ETH dataset without data augmentation.}
	\label{size_map}
\end{figure}

\noindent\textit{4.7.3  The size of probability trajectory}

The probability trajectory is two-dimensional, so the suitable size is necessary. We set up the comparison experiment, and the size is 80*80, 100*100, 150*150, 200*200. The results are shown in Fig.~\ref{size_map}. From the figure, we can find that when the size is 100*100, the predicted result is the best. Too large or too small will cause decreases in prediction accuracy. We sampled from the ground truth and found that as the size increases, the sampling error also increases. The sampling error may be the reason for the decrease of prediction accuracy. On the contrary, as the size decreases, the model is not capable to modeling large enough amounts of data.

\section{Conclusions}\label{sec:conclusions}

In this paper, we propose the concept of probability trajectory, which has more advantages in representing the randomness of trajectories, and explore a new trajectory prediction method based on it. To encode social interaction features, we introduce ConvLSTM, a sequence to sequence prediction model, which has the ability to model spatiotemporal characteristics. Experiments on public datasets show the effectiveness of our method. Although it is not state-of-the-art in all datasets, our method is simple and has great potential. In addition, our current work does not incorporate physical surrounding, but it is obvious that adding physical scenarios to our model is straightforward and convenient, and this is the direction of our future work.

\bibliographystyle{unsrt}
\bibliography{cvm21}

\end{document}